\documentclass[11pt]{article}

\usepackage{acl}

\usepackage{times}
\usepackage{latexsym}
\usepackage{amsmath}
\usepackage{longtable}
\usepackage{enumitem}
\usepackage{float}
\usepackage{booktabs}
\usepackage{tabularx}
\usepackage{array}
\usepackage{cuted}
\usepackage{capt-of}
\usepackage{tikz}
\usepackage{graphicx}
\usetikzlibrary{arrows.meta,calc}
\usetikzlibrary{arrows.meta,decorations.pathreplacing}
\usepackage[T1]{fontenc}

\usepackage[utf8]{inputenc}

\usepackage{microtype}

\usepackage{inconsolata}

\usepackage{graphicx}

\title{Can Generative AI Automate Data Extraction for Meta-Analysis? A Case Study on Intercropping Research}

\author{
  \textbf{Zehao Lu}\textsuperscript{1,*},
  \textbf{Xingguo Xiong}\textsuperscript{2,*},
  \textbf{Wopke van der Werf}\textsuperscript{1},\\
  \textbf{Thijs L. van der Plas}\textsuperscript{1},
  \textbf{Ioannis N. Athanasiadis}\textsuperscript{1}
  \\
  \textsuperscript{1}Wageningen University \& Research,
  Wageningen, the Netherlands
  \\
  \textsuperscript{2}Zhejiang Academy of Agricultural Sciences
  \\
  \small{
    \texttt{\{zehao.lu,wopke.vanderwerf,thijs.vanderplas,%
    ioannis.athanasiadis\}@wur.nl},
    \texttt{xiongxg@zaas.ac.cn}
  }
  \\
  \textsuperscript{*}\small Equal contribution.
}

\begin{document}
\maketitle
\begin{abstract}
Meta-analysis is the synthesis of information from multiple sources to arrive at an overarching conclusion. There is a large need for meta-analysis in agricultural research to synthesize what is known and analyze overarching patterns. Extracting data from published literature is, however, labor-intensive, time-consuming, and tedious, and is impeded by a lack of standardization in research design, units of measurement, and terminology. These challenges are particularly evident in the domain of crop species mixtures, also called intercropping. With the growing capabilities of LLMs, many recent attempts have focused on building systems and tools to automate data collection, yet rigorous assessment against human-labeled ground truth is often missing. In this research, we evaluate three LLM-based approaches---direct zero-shot prompting, a staged workflow, and a multi-agent system---with six open-weight models to extract data from the intercropping literature. The results are evaluated against the manually curated ground truth and through a downstream statistical analysis. Overall, direct zero-shot prompting is the strongest and most consistent approach, achieving the highest mean similarity-adjusted F1 of 0.577, although none of the approaches is close to fully accurate. In the downstream analysis, most model--approach combinations recover the direction of the relationship between the predictor and outcome variables, but do not estimate its magnitude accurately. \footnote{Code, prompts, and evaluation scripts are available at
\url{https://github.com/com3dian/meta_analysis_agents}.}
\end{abstract}
\section{Introduction}

Modern agriculture faces the dual challenge of increasing food production for a growing global population while adapting to climate change \cite{doi:10.1126/science.1185383}. Meeting these demands requires a transition from conventional monoculture toward more diversified cropping systems \cite{tamburini2020agricultural}. Intercropping is one such system, in which two or more crop species are grown together to improve resource-use efficiency and ecological resilience \cite{brooker2015improving}. However, its benefits vary across crop combinations, pedoclimatic conditions, and management practices such as sowing dates, planting densities, and fertilizer inputs \cite{antle2017next,grassini2015good}. Evidence from different environments must therefore be synthesized to identify suitable crop combinations and management practices.

Meta-analysis systematically combines quantitative evidence from published studies to assess outcomes across diverse conditions \cite{gurevitch2018meta}. This generally involves identifying eligible studies, extracting comparable outcomes and their uncertainty, and synthesizing the evidence statistically. In agriculture, this evidence can help evaluate management strategies and provide empirical data for models exploring production and climate-adaptation scenarios \cite{antle2017next,HOLZWORTH2014327}. Meta-regression extends this approach by examining whether study-level factors explain variation in outcomes across studies. This makes it necessary to extract not only the reported outcomes, but also the context in which they were observed. Because crop responses depend strongly on local soil, weather, and management conditions, a reported yield value is meaningful only when its experimental context is retained \cite{antle2017next,grassini2015good}. Meta-analysis in agricultural sciences therefore requires structured datasets that preserve not only numerical measurements, but also experimental designs, treatment comparisons, environmental conditions, and measurement units.

Constructing these datasets is particularly difficult for intercropping research, due to the lack of standardization and widely-accepted protocols, variations in equipment, and variability of metrics reported in each local study. A single trial may report dozens of related measurements across intercropped plots and their corresponding sole-crop controls. Studies also differ widely in their objectives, terminology, management conditions, outcomes, and units \cite{connolly2001information}. The information required for one experimental record is often scattered throughout a paper: trial designs may be described in the text, numerical results reported in multi-level tables, and units or exceptions provided only in captions or footnotes. Although human experts can interpret and connect this information, manual curation requires substantial time and effort, creating a major bottleneck for large-scale evidence synthesis.

Earlier attempts to automate evidence synthesis relied mainly on task-specific NLP pipelines, with separate components for document filtering, entity recognition, and relation extraction \cite{callaghan2021machine,sietsma2024machine,veena2023agroner,chebbi2024enhancing, rezayi2022agribert}. Although effective for well-defined subtasks, these pipelines can propagate errors between stages and often struggle to assemble information into complete experimental records. Many relevant relationships extend beyond individual sentences or paragraphs and require information to be connected across the full document \cite{jain-etal-2020-scirex}.

Large language models provide a more general approach: a single model can follow different extraction schemas, integrate broader context, and perform multiple extraction tasks \cite{kojima2023largelanguagemodelszeroshot, dagdelen2024structured}. This flexibility has led to growing interest in LLM-based scientific information extraction \cite{dagdelen2024structured,gupta2024data} and agentic designs such as ReAct, plan-and-execute, and AFlow \cite{yao2023reactsynergizingreasoningacting,erdogan2025planandactimprovingplanningagents,zhang2025aflowautomatingagenticworkflow}. Such designs may help decompose complex tasks, but additional stages can also introduce context loss and error propagation. Whether they improve extraction from complex agricultural studies remains an open question.

In this work, we investigate whether generative AI can reliably extract complex intercropping trial data. We compare a direct zero-shot approach, a staged workflow, and a multi-agent system across six LLMs. We evaluate their outputs against human-curated records and test their ability to reproduce a published meta-analysis. Our main contributions are:

\begin{enumerate}[
    leftmargin=1.6em,
    labelsep=0.4em,
    itemsep=0.2em,
    topsep=0.3em,
    parsep=0pt,
    partopsep=0pt
]
    \item We formulate the structuring of complex agricultural trial data as an information extraction task and establish an annotated evaluation benchmark.
    
    \item We compare three LLM-based approaches on multi-level trial structures and cross-table attribute binding.
    
    \item We assess how extraction errors affect the reproduction of published meta-analytic findings.
\end{enumerate}
\begin{figure*}[!t]
  \centering
  \resizebox{\textwidth}{!}{%
    \input{figures/example/example_fig.tex}%
  }
  \caption{Example of structured data extraction from Jensen's pea--barley intercropping paper \citep{jensen1996grain}, one of the 90 papers in our annotated corpus. Highlighted evidence from the experimental design, crop management, and yield table is linked to one ground-truth record. Asterisks mark values calculated from the reported table data.}
  \label{fig:extraction-example}
\end{figure*}
\section{Problem Statement}
\subsection{Structured Data Extraction}

Given the textual content of a research paper extracted from a PDF, our task is to identify information relevant to a predefined schema and organize it into a set of structured records. Information available only in images is outside the scope of the task. A paper may contain zero, one, or multiple records, and the information needed for one record may appear in different parts of the paper.

Each record represents one experimental comparison and contains a fixed set of fields defined by the schema. These fields may describe the study conditions, treatments, outcomes, and observed effects. Their standardization varies across disciplines, from protocol-based clinical trials to more heterogeneous agricultural field studies \citep{chan2013spirit,https://doi.org/10.2134/agronj2017.04.0215}. The task requires extracting the correct values while preserving the connections among values from the same comparison.

\subsection{Challenges Specific to Agricultural Studies}

Intercropping experiments often span multiple sites, seasons, and treatments, requiring each mixture to be matched with two sole-crop comparators under comparable management. This is difficult when fertilizer inputs or planting densities differ among treatments \cite{yu2015, liu2024nitrogen}. The production syndromes identified by \citet{Li2020} further show that crop composition, spatial design, management, timing, location, and yields must remain correctly linked. Outcomes may include crop-level yields, system-level yields, or indices such as the land equivalent ratio \cite{brooker2015improving}, often reported under nested table headers and in different units. Breaking these relationships can produce incorrectly linked or omitted records and distort downstream analysis.

\section{Data \& Experiment}

This section introduces the data used and the experimental setup of this study. We use three generative-AI-based approaches to repeat the data-extraction stage of \citet{yu2015}, hereafter referred to as the reference study. We then reproduce the subsequent analysis by fitting the same linear mixed-effects model specified in the original R analysis to the LLM-extracted outputs.

\subsection{Data}

The reference study manually collected 746 records from 100 intercropping papers: the 50 most highly cited papers in the analysed corpus and 50 selected at random. Only 90 of the 100 paper folders could be reliably mapped to the corresponding ground-truth studies, so we conduct the extraction experiment on these 90 papers.

The source papers were preprocessed using MinerU \cite{wang2026mineru2} and converted from PDF into Markdown documents, which serve as the textual input to the three extraction approaches.

The complete ground-truth dataset contains 135 variables, including bibliographic and administrative information outside the extraction task. We evaluate the predefined subset of 42 extraction variables without modifying their definitions or ground-truth values.

\subsection{Extraction Schema and Key Variables}

We use the field definitions provided by the author of the reference study to clarify what information is contained in each intercropping record. Definitions of all 42 extracted fields are provided in Appendix Table~\ref{tab:field-definitions}. Among these fields, Temporal niche differentiation (TND) and land-equivalent ratio (LER) are central to the reproduced analysis.

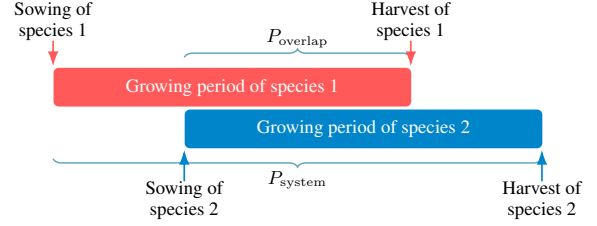
\begin{figure}[t]
\centering
\resizebox{\columnwidth}{!}{%
\begin{tikzpicture}[
    font=\small,
    event/.style={
        align=center,
        font=\small
    },
    period/.style={
        rounded corners=2.5pt,
        minimum height=0.65cm,
        align=center,
        text=white,
        font=\small
    }
]

\node[
    period,
    fill=red!65,
    minimum width=6.0cm,
    anchor=west
] (species1) at (1.0,1.55)
{Growing period of species 1};

\node[
    period,
    fill=cyan!65!blue,
    minimum width=6.0cm,
    anchor=west
] (species2) at (3.2,0.85)
{Growing period of species 2};

\node[event] at (1.0,2.65)
{Sowing of\\species 1};

\node[event] at (7.0,2.65)
{Harvest of\\species 1};

\node[event] at (3.2,-0.35)
{Sowing of\\species 2};

\node[event] at (9.2,-0.35)
{Harvest of\\species 2};

\draw[
    decorate,
    decoration={
        brace,
        mirror,
        amplitude=4pt
    },
    cyan!45!gray,
    line width=0.7pt
]
    (7.0,2.05) -- (3.2,2.05)
    node[
        midway,
        yshift=0.32cm,
        text=black
    ] {\(P_{\mathrm{overlap}}\)};

\draw[
    decorate,
    decoration={
        brace,
        mirror,
        amplitude=4pt
    },
    cyan!45!gray,
    line width=0.7pt
]
    (1.0,0.35) -- (9.2,0.35)
    node[
        midway,
        yshift=-0.35cm,
        text=black
    ] {\(P_{\mathrm{system}}\)};

\draw[
    red!65,
    line width=0.8pt,
    line cap=round,
    -{Latex[length=2.5mm,width=1.8mm]}
]
    (1.0,2.32) -- (1.0,1.90);

\draw[
    red!65,
    line width=0.8pt,
    line cap=round,
    -{Latex[length=2.5mm,width=1.8mm]}
]
    (7.0,2.32) -- (7.0,1.90);

\draw[
    cyan!65!blue,
    line width=0.8pt,
    line cap=round,
    -{Latex[length=2.5mm,width=1.8mm]}
]
    (3.2,-0.02) -- (3.2,0.50);

\draw[
    cyan!65!blue,
    line width=0.8pt,
    line cap=round,
    -{Latex[length=2.5mm,width=1.8mm]}
]
    (9.2,-0.02) -- (9.2,0.50);

\end{tikzpicture}%
}
\caption{Illustration of temporal niche differentiation. Adapted from \citet{yu2015}.}
\label{fig:tnd-definition}
\end{figure}

Temporal niche differentiation measures the difference between the growing periods of the two crop species in the case of relay intercropping. Relay intercropping is the combination with two species with differences in growing period, which overlap only partially.
\begin{equation}
    TND
    =
    1 -
    \frac{P_{\mathrm{overlap}}}
         {P_{\mathrm{system}}},
\end{equation}

where \(P_{\mathrm{overlap}}\) is the simultaneous growth period and \(P_{\mathrm{system}}\) spans the period from the sowing of the first-sown crop till the harvest of the last-harvested crop. A TND of 0 indicates complete overlap, while higher values indicate greater temporal separation (Figure~\ref{fig:tnd-definition}). The maximum value of TND is 1, indicating complete separation of growing periods.

The land-equivalent ratio measures how efficiently an intercropping system uses land relative to growing the two crop species separately. It is calculated as
\begin{equation}
    LER
    =
    \frac{Y_{\mathrm{IC1}}}{Y_{\mathrm{SC1}}}
    +
    \frac{Y_{\mathrm{IC2}}}{Y_{\mathrm{SC2}}},
\end{equation}

where \(Y_{\mathrm{IC1}}\) and \(Y_{\mathrm{IC2}}\) are species yields under intercropping, and \(Y_{\mathrm{SC1}}\) and \(Y_{\mathrm{SC2}}\) are their corresponding sole-crop yields. An LER above 1 indicates greater land-use efficiency under intercropping.

\subsubsection{Analysis in the Reference Study}

We reproduce the main analysis conducted in the reference study. This subsection describes that analysis and identifies the information required to reproduce it. The authors investigated the following research question:

\begin{quote}
\textit{Do crop combinations have a higher LER when their TND is larger, i.e. when they overlap (and compete) less?}
\end{quote}

To answer this question, the authors modelled the relationship between temporal niche differentiation (TND) and the land-equivalent ratio (LER). They used a linear mixed-effects regression with random intercepts for experiments nested within studies:
\begin{equation}
\begin{aligned}
LER_{ijk} ={}& \beta_0 + \beta_1 TND_{ijk} + u_{\mathrm{Study},i} \\
             &+ u_{\mathrm{Experiment},ij}
              + \epsilon_{ijk}.
\end{aligned}
\end{equation}
Here, \(\beta_0\) is the expected LER at \(\mathrm{TND}=0\), and \(\beta_1\) is the change in LER per unit increase in TND. The terms \(u_{\mathrm{Study},i}\) and \(u_{\mathrm{Experiment},ij}\) capture variation between studies and nested experiments, respectively, while \(\epsilon_{ijk}\) denotes residual variation.

\subsection{Experiment}

\subsubsection{Extraction Approaches}

We designed three LLM-based approaches: direct zero-shot prompting, a staged workflow, and a multi-agent system (MAS).

The prompts were developed based on referenced research and expert knowledge of intercropping meta-analysis. We tested different prompt versions during pilot runs and selected those that showed the best performance and stability. For the staged workflow, we followed previous work on separating evidence identification from record construction. In the first step, a labeller identifies relevant values and tags the supporting text. In the second step, an extractor uses the labeled text to construct the final records.

During the prototyping of the agentic approach, we tested several designs, including ReAct \cite{yao2023reactsynergizingreasoningacting} and Reflexion \cite{shinn2023reflexionlanguageagentsverbal}. However, these repeated reasoning steps did not consistently improve the output and often accumulated errors. Reflexion showed a similar problem: even when previous error patterns were provided, the system did not reliably avoid them. We therefore selected a plan-and-execute design in which agents use tools to identify and label relevant evidence before producing and refining the records \cite{erdogan2025planandactimprovingplanningagents}. Further details about the agents, prompts, and tools are provided in the Appendix~\ref{app:config}.

\subsubsection{Language Model Choice}
Regarding the LLM choice, we used open-weight models available through the SURF research infrastructure. This choice was motivated by concerns about sharing unpublished scientific work with commercial LLM providers and the possibility that related results could be pursued or released before the original researchers are ready to publish \cite{buckmaster2026statement}. The six LLMs included \textsc{Qwen3.6-27B-FP8}, \textsc{Qwen3.6-35B-A3B-FP8}, \textsc{Qwen3.5-122B-A10B-NVFP4}, \textsc{Gemma-4-31B-it-NVFP4}, \textsc{Llama-70B}, and \textsc{GPT-OSS-120B}.

\section{Evaluation}

We evaluate the extraction approaches in two steps: comparing the extraction outputs against the ground truth using a rigorous evaluation pipeline, and reproducing the experiment conducted by human researchers in the original meta-analysis study using the extraction outputs. The first step aims to assess how well the LLM-based approaches agree with the human-annotated ground truth and how performance differs across models and methods. The second step examines whether the experiment can be reproduced using imperfect extraction outputs and how much the resulting findings differ from those reported by the human researchers.

\subsection{Evaluation Against Ground Truth}

We evaluate the outputs at the field level using comparison rules tailored to different data types. These rules were reviewed and agreed upon with the senior author of the reference study who was deeply involved in curation of the ground-truth dataset. We first introduce the types of extraction error and their corresponding metrics, and then describe the evaluation pipeline used to align records, apply the comparison rules, and aggregate the results.

\subsubsection{Error Types}
The output of the three approach (direct, workflow, MAS) are saved in \texttt{csv} format, as is the ground truth. We classify the mismatches between the predictions and ground truth into three types, each measured by a corresponding metric:

\begin{itemize}[
    leftmargin=1.6em,
    labelsep=0.4em,
    itemsep=0.2em,
    topsep=0.3em,
    parsep=0pt,
    partopsep=0pt
]
    \item \textbf{Incorrectness}: Extracted values disagree with the corresponding ground-truth values in matched field values. Agreement is measured by the \textit{similarity score}, which ranges from 0 to 1 and is averaged over fields where both records contain a value.
    \item \textbf{Hallucination}: A predicted field contains a value, but its ground-truth counterpart is missing. Populated fields in unmatched predicted records also count as unsupported extractions. The \textit{factual rate} measures the proportion of populated predicted fields that have a populated ground-truth counterpart
    \item \textbf{Incompleteness}: A ground-truth field contains a value, but its predicted counterpart is missing. Populated fields in unmatched ground-truth records also count as omissions. The \textit{completeness rate} measures the proportion of populated ground-truth fields that have a populated predicted counterpart.
\end{itemize}

\subsubsection{Evaluation Metrics}
\textbf{Similarity Score} We define a field-level similarity function that compares an extracted value with its corresponding ground-truth value and returns a score between 0 and 1. A score of 1 indicates full agreement under the applicable comparison rules, while 0 indicates no match. Intermediate scores represent partial agreement. The comparison depends on the field type, as follows:
\begin{itemize}[
    leftmargin=1.5em,
    labelsep=0.4em,
    itemsep=0.15em,
    topsep=0.25em,
    parsep=0pt,
    partopsep=0pt
]
    \item \textbf{Text and categorical values}: After lowercasing and removing whitespace, we compute character-level ROUGE-L F1 \cite{lin-2004-rouge}. Identical normalized strings receive 1, while partial overlap receives proportional credit.

    \item \textbf{Numerical values}: Values receive 1 when their relative difference is at most 2\%; an exact match is required when the ground truth is zero. Values that match only after removing decimal points receive 0.5, reflecting a possible scale or unit-conversion error. All other mismatches receive 0.

    \item \textbf{Numerical values with units}: Compatible units are normalized and converted to the ground-truth unit before applying the 2\% tolerance. If either unit is missing, we use the unit-free numerical comparison; incompatible or invalid quantity comparisons receive 0.

    \item \textbf{Dates}: We normalize calendar and Excel serial dates and sequentially compare full dates, partial month--day or month--year dates, relative days or ranges, and years. Applicable rules return 1 or 0; otherwise, we use character-level ROUGE-L F1.

    \item \textbf{Boolean and missing values}: Equivalent Boolean forms, such as \textit{Yes}, \textit{true}, and 1, are normalized before exact comparison. A comparison receives 0 if either value is missing, including when both are missing.
\end{itemize}

\textbf{Factual Rate} To assess hallucination error, we define the factual rate as the proportion of non-empty predicted fields that have a non-empty ground-truth counterpart after record alignment. This metric measures the presence of a corresponding value, regardless of whether the predicted and ground-truth values agree.

\textbf{Completeness rate} To assess omissions, we define the completeness rate as the proportion of non-empty ground-truth fields that have a non-empty predicted counterpart after record alignment. This metric measures how much of the ground-truth information is covered by the extraction, regardless of whether the values agree.

\paragraph{Combined Evaluation Score} The three metrics capture complementary aspects of extraction quality. The similarity score $S$ measures the correctness of aligned values, the factual rate $F$ penalizes unsupported predictions, and the completeness rate $C$ penalizes omitted information. To summarize these dimensions in a single score, we first calculate a similarity-adjusted factual rate:
\begin{equation}
    F_{\mathrm{adj}} = S \times F,
\end{equation}

where \(S\) is the similarity score and \(F\) is the factual rate. We then combine \(F_{\mathrm{adj}}\) with the completeness rate \(C\) using their harmonic mean:
\begin{equation}
    E_{\mathrm{combined}}
    =
    \frac{2 F_{\mathrm{adj}} C}
         {F_{\mathrm{adj}} + C}
    =
    \frac{2SFC}{SF+C}.
\end{equation}

The resulting score ranges from 0 to 1, with higher values indicating better overall extraction quality. Multiplying similarity by the factual rate ensures that a populated predicted field contributes fully only when its value is also correct. The harmonic mean follows the principle of the F-measure and produces a high combined score only when both the similarity-adjusted factual rate and completeness rate are high \citep{schutze2008introduction}. We refer to this metric as the \emph{similarity-adjusted F1 score}.

\subsubsection{Evaluation pipeline}
For each approach, we compare the extracted records with the ground-truth records from the same paper. The evaluation includes papers with an output CSV from at least one approach; missing outputs from other approaches are treated as empty sets of records.

We first compare every predicted record with every ground-truth record from the same paper. For each pair, we calculate record-level similarity as the arithmetic mean of the evaluated field scores. To account for interchangeable crop ordering, we retain the higher-scoring original or swapped crop-1/crop-2 assignment.

We sort all candidate record pairs by descending similarity and perform greedy one-to-one matching. A pair is accepted only if neither record has already been matched. No minimum similarity score is required.

After matching, each non-empty field in an unmatched predicted record is counted as an unsupported extraction, while each non-empty field in an unmatched ground-truth record is counted as an omission. For matched records, we evaluate field presence and agreement using the retained crop orientation. We calculate paper-level similarity, factual rate, and completeness rate, then average these metrics across papers for each approach.

\subsection{Reproducing the Reference Analysis} \label{sec:reproducing}

Because the extracted datasets are imperfect, we investigate whether they nevertheless support the same or a similar scientific conclusion as the reference study. Automating meta-analysis should support not only accurate extraction but also the complete process from the literature to a final conclusion. We therefore examine how extraction errors propagate into the downstream analysis and whether they alter its findings.

For each model and extraction approach, we first construct the variables required for the regression. When a usable LER value is reported in an extracted record, we use that value directly. When LER is unavailable, we calculate it from the four required sole-crop and intercrop yields:
\begin{equation}
    LER =
    \frac{Y_{\mathrm{IC1}}}{Y_{\mathrm{SC1}}}
    +
    \frac{Y_{\mathrm{IC2}}}{Y_{\mathrm{SC2}}}.
\end{equation}
Records without either a reported LER or all four required yields are excluded. We similarly use the extracted TND when available or calculate it from the crops’ sowing and harvest dates. Only records for which both LER and TND can be obtained are retained.

Within each study, the retained records are matched one-to-one with the reference records using experimental-design information only. LER, TND, yields, and dates are not used during this matching. The matched ground-truth Experimental ID is used solely to recover the nested grouping structure required by the statistical model; no ground-truth outcome values are introduced into the reproduced analysis.

We apply the linear mixed-effects model described in the previous section, with LER as the response, TND as the fixed-effect predictor, and random intercepts for experiments nested within studies, fitting it separately for each model and extraction approach.

\section{Results}

We applied all three extraction approaches with each LLM to the same 90 papers. We counted runs that produced a parsable table with at least one record. Individual LLM requests had a three-minute timeout, although multi-stage runs could take longer. Failures included API or parsing errors, empty tables, and missing outputs. Table~\ref{tab:run-completion} summarizes the resulting coverage.

\begin{table}[t]
\centering
{\small
\setlength{\tabcolsep}{2.5pt}
\renewcommand{\arraystretch}{1.08}
\begin{tabular}{@{}lccc@{}}
\toprule
\textbf{LLM} & \textbf{Direct} & \textbf{Workflow} & \textbf{MAS} \\
\midrule
\textsc{Qwen3.6-27B}     & 48/0/42 & 41/5/44 & 50/12/28 \\
\textsc{Qwen3.6-35B-A3B} & 86/0/4  & 76/0/14 & 57/29/4  \\
\textsc{Gemma-4-31B}     & 69/1/20 & 65/0/25 & 74/6/10  \\
\textsc{Qwen3.5-122B}    & 71/0/19 & 68/0/22 & 46/34/10 \\
\textsc{Llama-70B}       & 74/1/15 & 73/3/14 & 55/22/13 \\
\textsc{GPT-OSS-120B}    & 75/1/14 & 73/2/15 & 68/13/9  \\
\bottomrule
\end{tabular}
}
\caption{Run outcomes across the 90 target papers. Each cell reports
non-empty/empty/missing outputs. Direct denotes the direct LLM
approach, Workflow the staged workflow, and MAS the multi-agent system.}
\label{tab:run-completion}
\end{table}

For each model, the evaluation includes papers for which at least one approach produced an output CSV. Within this set, a missing or empty output from a particular approach is treated as an empty prediction. Consequently, all non-empty ground-truth fields for that approach are counted as omissions.

\subsection{Field-Level Extraction Results}

\begin{figure}[H]
  \centering
  \includegraphics[width=\columnwidth]{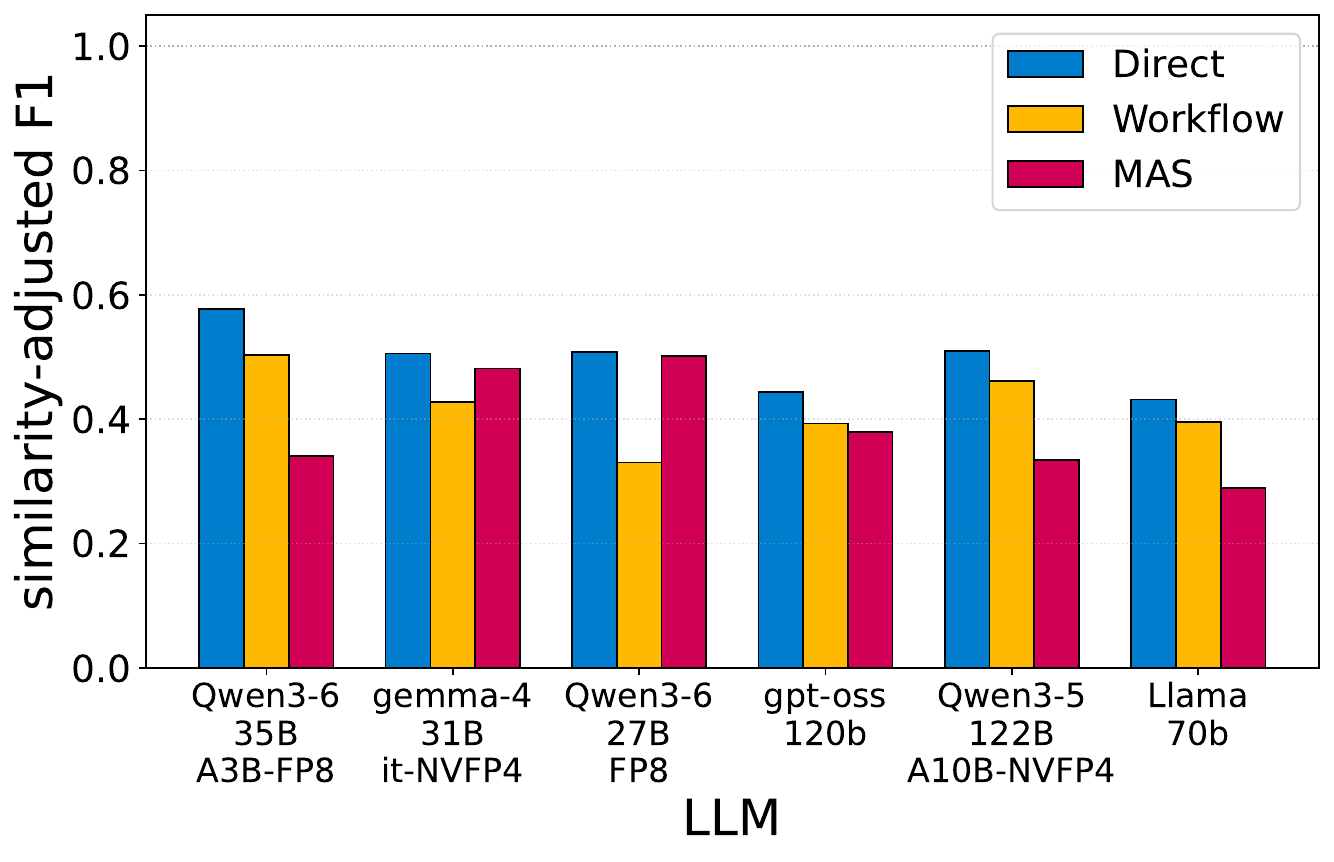}
  \caption{Mean similarity-adjusted F1 by LLM and extraction approach. Papers with output from at least one approach are included; missing or empty outputs are treated as empty predictions.}
  \label{fig:result_1}
\end{figure}

We applied the evaluation pipeline described above to the outputs of each LLM and extraction approach. Figure~\ref{fig:result_1} presents the mean similarity-adjusted F1 score for all papers included in the evaluation of each model. A paper is included when at least one of the three approaches produces an output CSV. For the included papers, a missing or empty output from an approach is evaluated as an empty prediction.

Figure~\ref{fig:result_1} shows that the direct LLM approach achieves the highest score for all six models. Performance also varies across LLMs, but the most consistent pattern is the difference between extraction approaches: increasing method complexity does not improve extraction quality. Contrary to our initial expectation, the staged workflow and multi-agent system generally perform worse than the direct approach. 

\begin{figure}[H]
  \centering
  \includegraphics[width=\columnwidth]{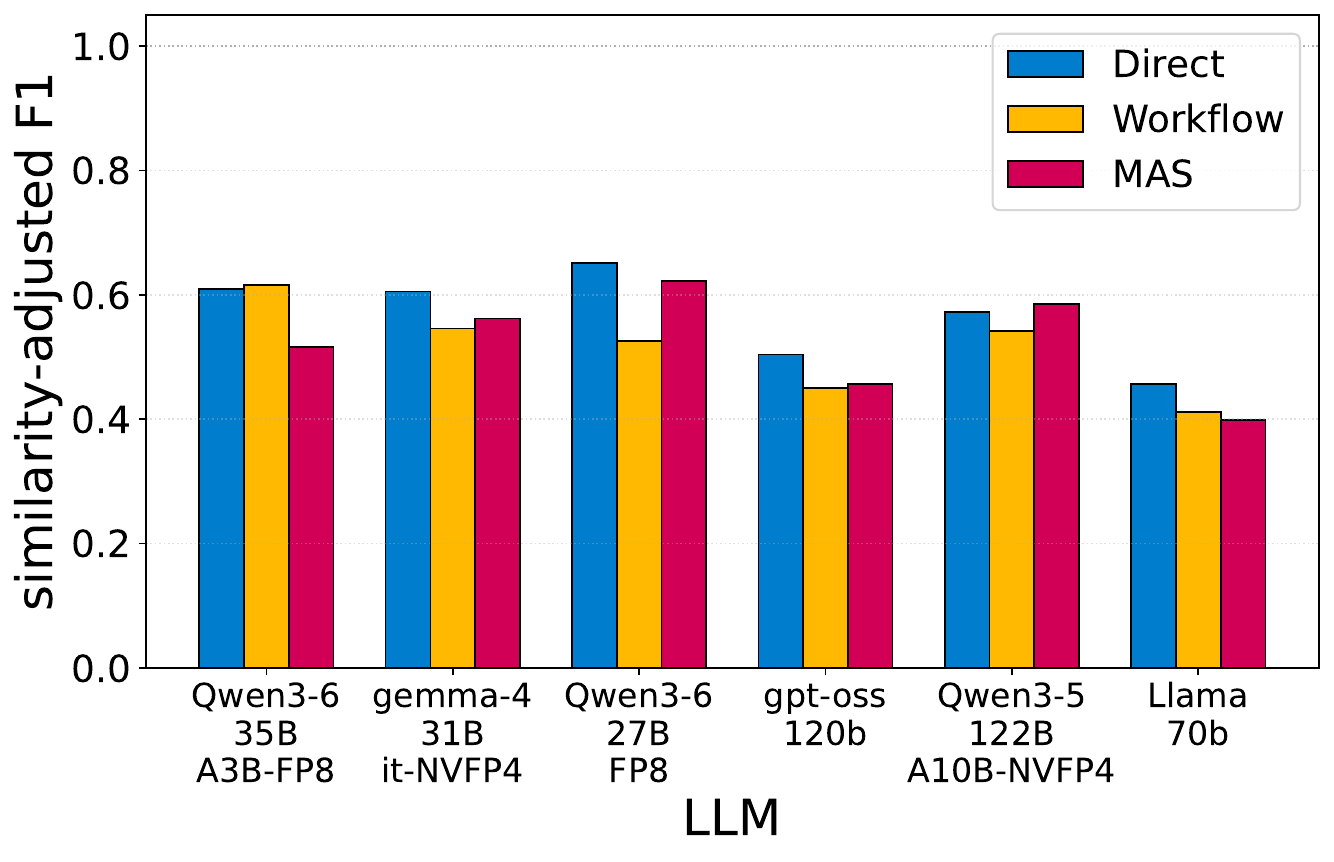}
  \caption{Mean similarity-adjusted F1 scores restricted, for each LLM,
  to papers for which all three extraction approaches produced at least
  one structured record.}
  \label{fig:result_2}
\end{figure}

This comparison is partly influenced by differences in run coverage. As
shown in Table~\ref{tab:run-completion}, the direct approach produces more
non-empty outputs overall. We therefore conduct an additional comparison
restricted, for each model, to papers for which all three approaches produce
at least one structured record. Figure~\ref{fig:result_2} presents the results
on this common-output subset.

The restricted comparison leads to a similar overall conclusion. The direct
approach obtains the highest score for four of the six models. The workflow
is slightly better for Qwen3.6-35B, while the multi-agent system is slightly
better for Qwen3.5-122B. Thus, although the ranking is not identical for every model, the results
provide no consistent evidence that increasing method complexity improves
extraction quality.

\subsection{Downstream Reproduction Results}

We reproduce the reference analysis by fitting the same linear mixed-effects model specification separately to the data produced by each LLM and extraction method, as described in Section~\ref{sec:reproducing}. Because extraction was conducted on 90 papers, we use the model fitted to the corresponding 90 ground-truth studies as the primary reference. This linear regression model estimates a positive TND effect of $\beta=0.290$ with a 95\% confidence interval of $[0.166, 0.413]$. We also show the effect reported for the complete reference dataset of 100 studies ($\beta\approx0.211$).

\begin{figure}[t]
    \centering
    \includegraphics[width=\columnwidth]{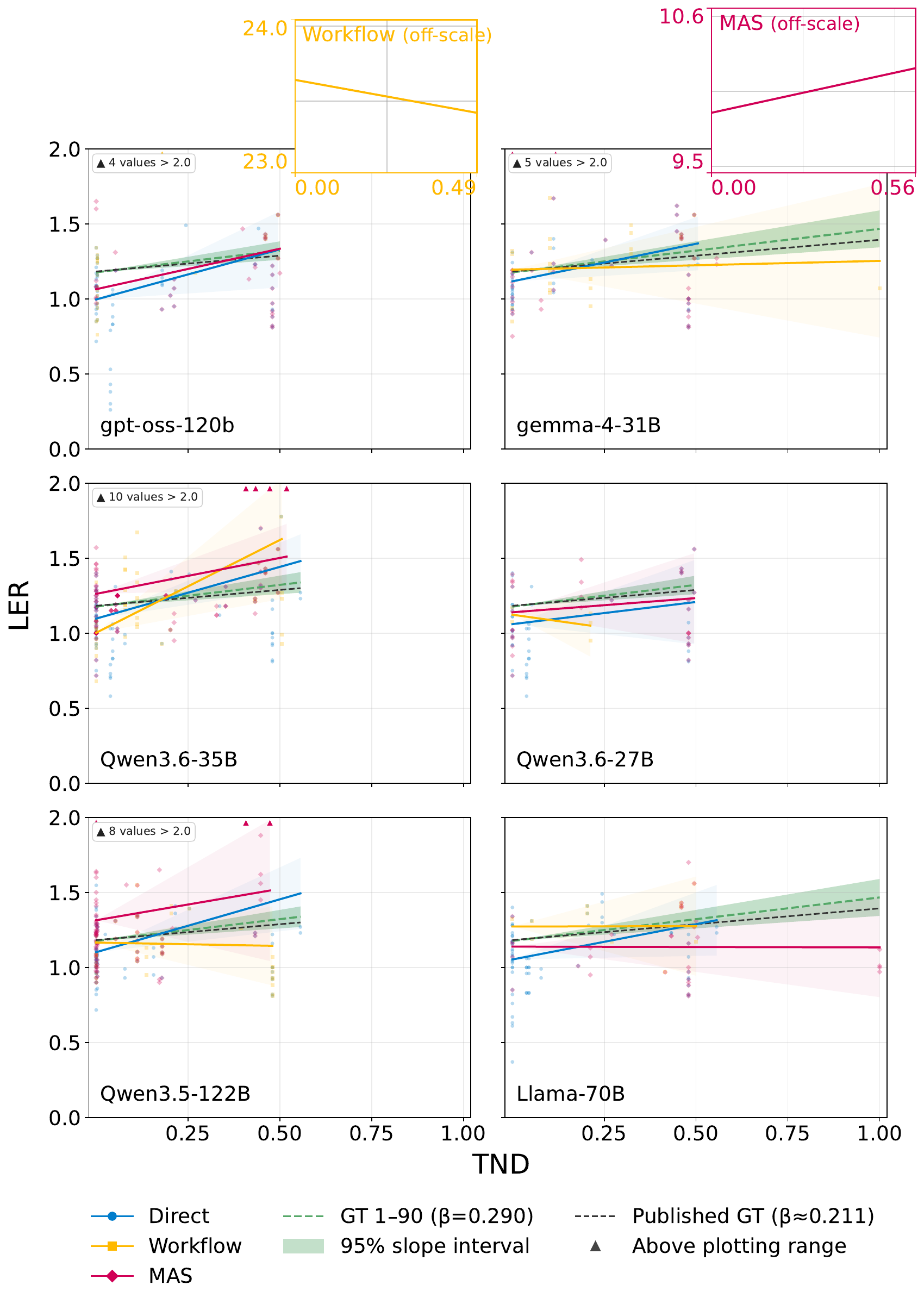}
    \caption{Estimated relationship between TND and LER for each LLM and extraction approach, compared with the ground-truth baselines. Insets show fits outside the main plotting range.}
    \label{fig:ler-tnd-reproduction}
\end{figure}

Figure~\ref{fig:ler-tnd-reproduction} shows the reproduced relationships. The direct approach produces a positive TND coefficient for all six LLMs, and five of these six estimates are significantly greater than zero. The Qwen3.6-35B model provides the most consistent qualitative reproduction: all three extraction methods produce significant positive coefficients. However, their estimated magnitudes are larger than the 90-study ground-truth coefficient. In contrast, the direct Qwen3.6-27B estimate ($\beta=0.295$) is numerically closest to the ground truth, but its confidence interval includes zero. The GPT-OSS multi-agent fit produces a coefficient estimate but no valid standard error or confidence interval.

Overall, the automatically extracted data often recover the direction of the
relationship, particularly when using the direct approach, but they do not
reliably reproduce its magnitude or statistical certainty. The results
therefore support the use of generative-AI extraction for qualitative
exploration, but not as a substitute for human-curated data when precise
quantitative conclusions are required.

Taken together, the field-level evaluation and reproduction experiment provide no consistent evidence that increasing method complexity improves extraction quality or downstream analytical reliability. We discuss possible explanations for this result in Section~\ref{subsec:complexity-trap}.

\section{Discussion \& Conclusion}

Compared with the ground truth, direct zero-shot extraction was generally the strongest approach, although none of the approaches was fully accurate. It consistently recovered the positive relationship between TND and LER, but not the magnitude of the reference effect.

\subsection{Complexity Trap} \label{subsec:complexity-trap}
We manually examined the outputs for a fixed-seed sample of eight papers to better understand why the structured pipelines (workflow and MAS) performed less well on average. The two-step workflow sometimes lost context when papers were reduced to tagged evidence blocks, while MAS could carry mistakes from its initial direct extraction into later steps. In this sample, MAS produced eight records for a paper containing three ground-truth records, fifteen for a paper containing two, and no records for two other papers. Additional stages may help in some cases, but they also create more opportunities to omit, duplicate, or incorrectly associate evidence. The stronger overall performance of direct extraction may therefore be partly explained by its shorter processing path.

These observations align with reports of errors at agent handoffs \citep{lin2026agentaskmultiagentsystemsneed}, error amplification in agent systems \citep{kim2026sciencescalingagentsystems}, and mixed benefits of multi-agent decomposition on geometry benchmarks \citep{e-sobhani-etal-2026-multi}. Complex pipelines therefore depend on a good fit with the task and careful checking of intermediate outputs.

\subsection{Toward More Efficient Evidence Synthesis}
Extracting data for an intercropping meta-analysis is time-consuming. Papers may report several experiments across different sites or years, with multiple treatments in each experiment. Researchers must therefore identify the appropriate intercrop and sole-crop treatments and decide which comparisons are scientifically meaningful.

The process is also iterative because intercropping meta-analysis is not yet fully standardized and new response metrics continue to be developed and evaluated. Researchers often move repeatedly between data retrieval, analysis, and conceptualization before arriving at an extraction and analysis strategy that fits the available evidence. Experience from conducting the reference meta-analysis suggests that a small meta-analysis, such as an MSc thesis project, may take about six months. A larger meta-analysis covering 50--150 studies may require two to three years of a PhD project \cite{yu2015}. As experience in synthesizing intercropping research grows, there is an opportunity to standardize and streamline this process and make greater use of automated extraction.

Our results suggest that LLMs could reduce some of this manual work, but their outputs still require careful checking. Direct zero-shot extraction was generally the most reliable approach, although many records and fields remained missing, incorrect, or wrongly associated. This provides cautious support for earlier findings on LLM-based information extraction for evidence synthesis \citep{polak2024extracting}. In the downstream analysis, direct extraction recovered a positive relationship between TND and LER across all six models, but the estimated effects were generally larger than the reference effect and varied in precision.
These findings suggest that LLMs are currently more suitable for assisting researchers with data extraction than for fully automating the process without human verification.

LLMs may also reduce the time required for evidence synthesis by supporting study screening and full-text eligibility assessment, although their outputs must still be checked before drawing scientific conclusions \citep{DelgadoChaves2025,Taherinezhad2026,Wang2025}.

\subsection{Relevance Beyond Intercropping}
Intercropping studies are particularly difficult to conduct meta-analysis on because of their nested experimental structure and the complex relationships among treatments, crops, and values reported across tables. Similar challenges occur in other domains, such as clinical trials \citep{Nye2018} and materials-science datasets \citep{Zheng2023}.

The results from direct zero-shot prompting suggest that this approach may be transferred to other domains. However, its precision remains limited, and recovering the direction of a relationship appears easier than estimating its magnitude accurately. Domain-specific post-training may therefore be needed to improve performance.

\section*{Generative AI usage}

AI coding tools (Claude, Cursor, Codex) were used to co-write parts of the code base and for minor copy-editing of the manuscript. All code has been manually verified and the authors take full responsibility of all presented methods and results.

\section*{Acknowledgements}
This work was partially supported by the Horizon Europe project PHENET – Tools and Methods for Extended Plant Phenotyping and Envirotyping Services of European Research Infrastructures (Grant Agreement No. 101094587) and the Dutch Research Council (NWO) Large-Scale Research Infrastructures (LSRI) programme through the LTER-LIFE infrastructure (Grant No. 184.036.014).


\bibliography{custom}
\clearpage
\appendix
\onecolumn

\section{Schema Definition}
\label{sec:appendix-schema}

\small
\renewcommand{\arraystretch}{1.08}

\begin{longtable}{
    @{}
    >{\raggedright\arraybackslash}p{0.22\textwidth}
    >{\raggedright\arraybackslash}p{0.75\textwidth}
    @{}
}
\caption{Definitions of the 42 fields in an extracted intercropping record.}
\label{tab:field-definitions}\\
\toprule
\textbf{Field} & \textbf{Definition} \\
\midrule
\endfirsthead

\multicolumn{2}{l}{\small\textit{Table \thetable{} continued}}\\
\toprule
\textbf{Field} & \textbf{Definition} \\
\midrule
\endhead

\midrule
\multicolumn{2}{r}{\small\textit{Continued on the next page}}\\
\endfoot

\bottomrule
\endlastfoot

Year of data &
Year or years in which the experimental data were collected. \\

Duration of experiment &
Total duration of the experiment, expressed in days, years, or growing seasons. \\

Experimental design &
Experimental layout used in the study, such as a randomized complete block design. \\

Sowing date 1 &
Date on which the first crop species in the intercropping system was sown. \\

Sowing date 2 &
Date on which the second crop species in the intercropping system was sown. \\

Harvest date 1 &
Date on which the first crop species was harvested. \\

Harvest date 2 &
Date on which the second crop species was harvested. \\

TND &
Temporal niche differentiation index explicitly reported in the paper. \\

Lat &
Latitude of the experimental site in decimal degrees. \\

Lon &
Longitude of the experimental site in decimal degrees. \\

Continent &
Continent in which the experimental site is located. \\

Crop species 1 &
Common or scientific name of the first crop species in the intercropping system. \\

Crop species 2 &
Common or scientific name of the second crop species in the intercropping system. \\

Crop type 1 &
Agronomic functional type of the first crop, such as cereal, legume, oilseed, or root crop. \\

Crop type 2 &
Agronomic functional type of the second crop, such as cereal, legume, oilseed, or root crop. \\

Fodder crop 1 &
Boolean indicator of whether the first crop is grown for fodder, forage, or biomass rather than for grain or seed. \\

Fodder crop 2 &
Boolean indicator of whether the second crop is grown for fodder, forage, or biomass rather than for grain or seed. \\

Intercropping pattern &
Spatial arrangement of the crops, classified as row, strip, or mixed intercropping. \\

Density ic 1 &
Plant density of the first crop species in the intercropping treatment. \\

Density ic 2 &
Plant density of the second crop species in the intercropping treatment. \\

Density sc 1 &
Plant density of the first crop species when grown as a sole crop. \\

Density sc 2 &
Plant density of the second crop species when grown as a sole crop. \\

RDT &
Relative density total of the intercropping system when explicitly reported. \\

N input SC1 &
Amount of nitrogen fertilizer applied to the sole-crop treatment of the first crop species. \\

N input SC2 &
Amount of nitrogen fertilizer applied to the sole-crop treatment of the second crop species. \\

N input IC1 &
Amount of nitrogen fertilizer attributed to the first crop species in the intercropping treatment. \\

N input IC2 &
Amount of nitrogen fertilizer attributed to the second crop species in the intercropping treatment. \\

N total in IC &
Total amount of nitrogen fertilizer applied to the intercropping treatment. \\

N Unit &
Unit used to report nitrogen fertilizer inputs, such as kg N ha$^{-1}$. \\

Replications SC1 &
Number of replicates for the sole-crop treatment of the first crop species. \\

Replications SC2 &
Number of replicates for the sole-crop treatment of the second crop species. \\

Replications IC1 &
Number of replicates for the first crop species in the intercropping treatment. \\

Replications IC2 &
Number of replicates for the second crop species in the intercropping treatment. \\

Data source &
Location of the extracted values in the publication, such as a table, figure, or supplementary material. \\

unified yield sc 1 &
Yield of the first crop species when grown as a sole crop. Grain yield is used for grain crops, while total dry-matter or biomass yield is used for fodder crops. \\

unified yield sc 2 &
Yield of the second crop species when grown as a sole crop. Grain yield is used for grain crops, while total dry-matter or biomass yield is used for fodder crops. \\

unified yield ic 1 &
Intercrop yield of the first crop species. For multi-cut forage crops, this is the cumulative seasonal yield across cuts. \\

unified yield ic 2 &
Intercrop yield of the second crop species. For multi-cut forage crops, this is the cumulative seasonal yield across cuts. \\

Yield unit &
Common unit used for the four sole-crop and intercrop yield fields, such as t ha$^{-1}$, kg ha$^{-1}$, or g m$^{-2}$. \\

PLER 1 &
Partial land-equivalent ratio of the first crop species when explicitly reported. \\

PLER 2 &
Partial land-equivalent ratio of the second crop species when explicitly reported. \\

LER &
Total land-equivalent ratio of the intercropping treatment when explicitly reported. \\

\end{longtable}

\normalsize
\twocolumn
\section{Prompts and Agent Configurations}
\label{app:config}

\subsection{Direct Zero-Shot Prompt}
\label{app:direct-zero-shot-prompt}

The direct approach used the following prompt template. At runtime,
\texttt{[SCHEMA]} was replaced by the complete 42-field schema and
\texttt{[PAPER CONTENT]} by the paper text. The model returned structured
output conforming to the same schema.

\begin{quote}
\small

You are an expert agricultural data extraction specialist. Your task is to
extract measured crop yield information and related agronomic and contextual
variables from scientific research papers. Maximize recall: your output list
should include every distinct experiment-level or table-level data row
supported by the paper, rather than a summary.

\textbf{Step 1: Anchor identification}

Scan the entire content, including paragraphs, all tables row by row,
captions, footnotes, and supplements. Locate every sentence or cell reporting
measured yields, densities, inputs, or other schema fields. Ignore model
evaluation metrics, yield gaps, and simulated outputs.

\textbf{Step 2: Contextual reasoning}

For each anchor, gather the associated context, including year, treatment,
fertilization level, cropping system, species, and block or plot when
reported.

\textbf{Step 3: Completeness, evidence, and confidence}

If a field is missing from a record, search the following locations before
leaving it null:

\begin{itemize}
    \setlength{\itemsep}{0pt}
    \setlength{\parskip}{0pt}
    \item Year, season, or duration: abstract, methods, table-row labels,
    and column headers.
    \item Location or coordinates: site description, methods, and tables.
    \item Crops, species, or intercropping pattern: abstract, methods, and
    table or figure labels.
    \item Treatments: experimental design, methods, and table columns.
    \item Densities, inputs, and nutrients: methods and numerical table cells.
    \item Yield columns: map sole-crop and intercrop yields to the
    corresponding schema fields for each row.
\end{itemize}

Assign confidence as high, medium, or low, but still output the row when its
values are supported by the paper.

\textbf{Step 4: Record construction}

Build one schema record for each distinct experimental observation:

\begin{itemize}
    \setlength{\itemsep}{0pt}
    \setlength{\parskip}{0pt}
    \item Treat each row of a main results table, such as a distinct
    year $\times$ treatment $\times$ system combination, as a separate record.
    \item When a table contains separate columns for the two species and their
    intercrop yields, fill all applicable yield fields for that row.
    \item For factorial or split-plot experiments, create a separate record
    for each factor combination with its own numerical result.
    \item Keep the original units and do not convert them. Use \texttt{null}
    only when a field is not reported for that row.
\end{itemize}

Remove duplicates only when two records have the same year, treatment level,
cropping system, and species context. When uncertain, keep both records.

\textbf{Target data}

Include field-measured yields, dry matter, densities, nutrient applications,
and other values defined by the schema. Exclude model evaluation metrics,
predictions, and correlation-only statistics.

Extract all rows supported by the main results tables. The number of records
should generally be similar to, or greater than, the number of distinct table
rows and relevant treatment levels. For intercropping tables, produce
separate records or fill the sole-crop and intercrop yield columns for each
row. Dry-matter measurements reported in units such as
$\mathrm{g\,m^{-2}}$ and $\mathrm{kg\,ha^{-1}}$ are valid.

\textbf{Meta-analytic schema}

\texttt{[SCHEMA]}

Analyze the following paper and extract all records according to the schema:

\texttt{[PAPER CONTENT]}

Extract all records from the paper following the schema. Prefer too many
distinct records over too few.

\end{quote}

\subsection{Staged Workflow Prompts}
\label{app:workflow-prompts}

The staged workflow consisted of two steps. The first step selected and tagged
schema-relevant evidence from the paper. The second step used the tagged text to
produce the structured records.

\subsubsection{Step 1: Evidence Labelling}

The following prompt was used for the labelling step. \texttt{[SCHEMA FIELDS]}
was replaced by the names and descriptions of all schema fields, and
\texttt{[PAPER CONTENT]} was replaced by the paper text.

\begin{quote}
\small
You are a document labeller for downstream structured extraction.

\textbf{Task:} Produce a single field \texttt{labeled\_document} containing
only the paragraphs, table blocks, captions, and footnotes that hold
schema-relevant evidence. Drop all other text. Within each retained block, add
inline XML tags so that the next step can produce one output record per
experimental row.

\textbf{Schema fields:} Tag names must match these field names exactly:

\texttt{[SCHEMA FIELDS]}

\textbf{Keep:}
\begin{itemize}
    \item Yield, density, nutrient, experimental-design, crop, site, year, and
    treatment values.
    \item Results tables together with their captions and footnotes. Retain the
    complete table rather than a summary.
    \item Methods sentences reporting densities, fertilizer inputs,
    experimental design, sowing or harvest dates, or geographical coordinates.
    \item Any sentence containing a schema-relevant number or label.
\end{itemize}

\textbf{Drop:}
\begin{itemize}
    \item Sections without schema-relevant values, such as general background,
    acknowledgements, references, and unrelated discussion.
    \item Do not rewrite or paraphrase retained text. Copy the source wording
    and only add XML tags.
\end{itemize}

\textbf{Tagging rules:}
\begin{itemize}
    \item Wrap every schema-relevant number or label using
    \texttt{<FieldName>exact source text</FieldName>}.
    \item In tables, tag every cell that corresponds to a schema field,
    especially cells in individual yield, density, and input rows.
    \item Tag distinct years, treatments, and nitrogen levels so that records
    can be separated in the next step.
    \item Do not invent text; only wrap spans found in the paper.
    \item Preserve the table structure and separate retained blocks with a
    blank line.
\end{itemize}

\textbf{Input document:}

\texttt{[PAPER CONTENT]}

Return only one structured object containing
\texttt{labeled\_document}.
\end{quote}

\subsubsection{Step 2: Record Extraction}

The second step reused the direct zero-shot prompt in
Appendix~\ref{app:direct-zero-shot-prompt}, with the labelled output from the
first step supplied as the paper content. The following instruction was added
before that prompt:

\begin{quote}
\small
The paper below includes XML tags in the form
\texttt{<FieldName>verbatim text</FieldName>}, where the tag name matches a
schema field. Use these tags as evidence, but also examine every table row and
caption in the tagged text to maximize the number of complete records. The
absence of a tag does not necessarily mean that the corresponding information
is absent.
\end{quote}

The following completeness instructions were appended:

\begin{quote}
\small
\textbf{Output completeness:}
\begin{itemize}
    \item Produce as many records as the evidence supports, typically one
    record for each distinct row in the main results tables.
    \item For intercropping tables, populate all available yield columns for
    each row rather than combining the table into a small number of summary
    records.
    \item Do not merge rows that differ by year, treatment, or nitrogen level
    unless the schema explicitly requires an aggregate record.
    \item Examine the complete tagged document, including the methods, results,
    and all retained tables.
    \item Use \texttt{null} only when a field is genuinely unavailable after
    examining all relevant evidence.
    \item When uncertain between one combined record and several records,
    prefer separate records that preserve the structure of the source table.
\end{itemize}
\end{quote}

\subsection{Agents and Tools}
\label{app:mas-agents-tools}

The MAS used a planning agent followed by four execution agents. Table~\ref{tab:mas-agents}
summarizes their roles and tool access. Only the labeller used a callable tool;
the other agents operated on the document and intermediate outputs provided in
their context.
\begin{table}[H]
\centering
\small
\caption{Agents used in the MAS pipeline.}
\label{tab:mas-agents}
\renewcommand{\arraystretch}{1.08}

\begin{tabularx}{\columnwidth}{
    @{}
    >{\raggedright\arraybackslash}p{0.25\columnwidth}
    >{\raggedright\arraybackslash}X
    @{}
}
\toprule
\textbf{Agent} & \textbf{Description} \\
\midrule

Planner
& Creates the four-step execution plan and assigns each step to the corresponding agent. \\

Value identifier
& Scans the complete paper and identifies values associated with fields in the extraction schema. \\

Labeller
& Uses the \texttt{xml\_tag\_from\_field\_values} tool to insert field-specific XML tags around the identified values while preserving the original document. \\

Direct extractor
& Applies the direct zero-shot extraction procedure to the original paper and produces an initial set of structured records. \\

Record extractor
& Refines the initial records using the XML-labelled document, filling missing fields and correcting values when supported by the tagged evidence. \\

\bottomrule
\end{tabularx}
\end{table}
\end{document}